\documentclass{article}

\usepackage{microtype}
\usepackage{graphicx}
\usepackage{subfigure}
\usepackage{booktabs} 
\usepackage{listings}
\usepackage[]{algorithm2e}
\usepackage{multirow}
\usepackage{amsmath}

\usepackage{hyperref}

\usepackage{xcolor, soul}
\definecolor{og}{rgb}{0,0.6,0}

\usepackage[accepted]{mlsys2023}

\mlsystitlerunning{Learning to Parallelize with OpenMP by Augmented Heterogeneous AST Representation}

\begin{document}

\twocolumn[
\mlsystitle{Learning to Parallelize with OpenMP by Augmented Heterogeneous AST Representation}



\mlsyssetsymbol{equal}{*}

\begin{mlsysauthorlist}
\mlsysauthor{Le Chen}{isu}
\mlsysauthor{Quazi Ishtiaque Mahmud}{isu}
\mlsysauthor{Hung Phan}{isu}
\mlsysauthor{Nesreen K. Ahmed}{intel}
\mlsysauthor{Ali Jannesari}{isu}
\end{mlsysauthorlist}
\pagenumbering{roman}
\mlsysaffiliation{isu}{Department of Computer Science, Iowa State University, Ames, USA}
\mlsysaffiliation{intel}{Intel Labs, Santa Clara, CA, USA}

\mlsyscorrespondingauthor{Le Chen}{lechen@iastate.edu}
\mlsyscorrespondingauthor{Ali Jannesari}{Jannesar@iastate.edu}

\mlsyskeywords{code analysis, ML for systems}

\vskip 0.3in

\begin{abstract}
Detecting parallelizable code regions is a challenging task, even for experienced developers. Numerous recent studies have explored the use of machine learning for code analysis and program synthesis, including parallelization, in light of the success of machine learning in natural language processing. However, applying machine learning techniques to parallelism detection presents several challenges, such as the lack of an adequate dataset for training, an effective code representation with rich information, and a suitable machine learning model to learn the latent features of code for diverse analyses. To address these challenges, we propose a novel graph-based learning approach called Graph2Par that utilizes a heterogeneous augmented abstract syntax tree (Augmented-AST) representation for code. The proposed approach primarily focused on loop-level parallelization with OpenMP. Moreover, we create an OMP\_Serial dataset with 18598 parallelizable and 13972 non-parallelizable loops to train the machine learning models. Our results show that our proposed approach achieves the accuracy of parallelizable code region detection with 85\% accuracy and outperforms the state-of-the-art token-based machine learning approach. These results indicate that our approach is competitive with state-of-the-art tools and capable of handling loops with complex structures that other tools may overlook.
\end{abstract}

]



\printAffiliationsAndNotice{}  

\section{Introduction}
\label{intro}

\begin{figure*}[h]
    \centering
    \includegraphics[width=16cm]{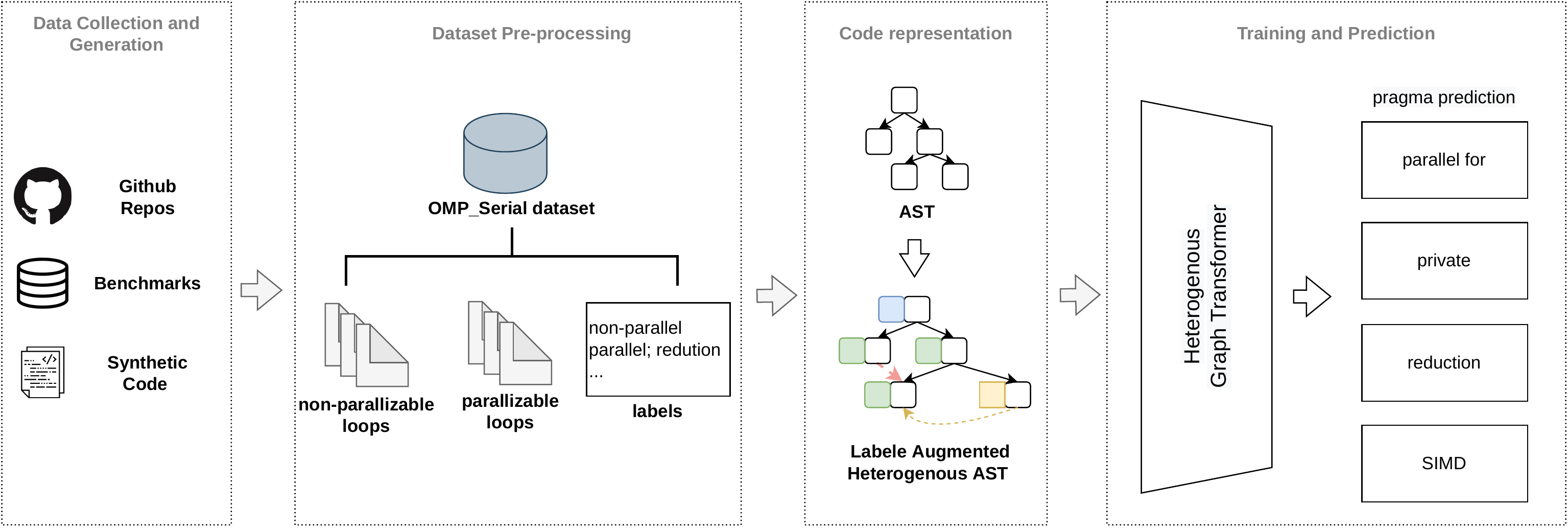}
    \caption{Proposed methodology. Data collection and generation: our dataset contains data from GitHub crawling, benchmark collection, and synthetic data generation. Data pre-processing: we extracted loops from codes with pre-processing steps, e.g., removing comments and blank lines. We also label the data according to the extracted pragma. Code representation: we generate the AST of each loop data and convert it to our proposed augmented heterogeneous AST. Training and Prediction: we feed our processed data and corresponding labels to the HGT model for 4 different downstream tasks.}
    \label{fig:pipeline}
\end{figure*}

The growing demand and popularity for multi-core hardware systems over the past few decades require developing highly-parallel programs to maximize performance. Numerous parallel programming models and frameworks \cite{chandra2001parallel, gabriel2004open, pheatt2008intel, bik2002automatic} have been created to facilitate the development of parallel code, but the developer's expertise in using these frameworks and familiarity with the codes are crucial to achieving better performance. Loop-level auto-parallelism helps developers in carrying out parallel tasks within the loops to speed up the process. Modern compilers typically detect the loop-level parallelism during compile time statically. This process is conservative and overlooks parallelism to ensure the correctness of the detected parallelism opportunities. On the other hand, dynamic auto-parallelism tools detect loop-level parallelism at runtime. The dynamic information captured after executing the programs improves the accuracy but has overhead issues. Moreover, the application of current auto-parallelization tools is constrained by requiring either compilation or execution of the programs for analysis. Therefore, a more practical way to auto-detect parallelism is required. 

Machine learning (ML) techniques are usually more feasible and cost-effective by redefining conventional software engineering problems as prediction problems. Many attempts have been made recently to use machine learning and Natural Language Processing (NLP) techniques in software engineering, from performance optimization and passes in compilers to solving complex problems such as malicious code detection, code placement on CPU or GPU, and performance prediction. Auto-parallelization with ML techniques is also conducted in recent studies. Chen et al.~\cite{chen2022multi} detect parallelism by training code static and dynamic information in a multi-view model. The code embedding in their work is an adaption of word2vec~\cite{mikolov2013efficient}, a now classic NLP technique.
Ben-nun et al. \cite{ben2018neural} introduce a Neural Code Comprehension (NCC) representation of code by using graph embeddings that are trained on unlabelled data before being used for simple code comprehension tasks.
Brauckmann et al.~\cite{brauckmann2020compiler} show that graph-embedding methods applied to Abstract Syntax Tree (AST) or Control Data Flow Graph (CDFG) are more efficient at downstream tasks than the state-of-the-art (NLP-inspired) methods, with better ability to generalize to never-seen-before examples. 

Despite their success, previous studies have shown common challenges in applying ML and NLP techniques in code analysis.  First, constructing relevant datasets is a major pain point when attempting to solve any problem using machine learning. Only a few public benchmarks for parallelization using OpenMP are applicable to the parallelism detection task. Second, code representation is crucial for machine learning models to comprehend programs. The intuitive solution is treating code as a natural language so NLP models can be applied directly \cite{dai2019transformer}. 
However, the context or token representation overlooks the code's structural information, which is crucial for parallelization analysis \cite{blume1994automatic, chen2022multi}. Finally, the performance of ML models varies across different tasks.


In this work, we propose to leverage state-of-the-art machine learning techniques to detect loop parallelism and suggest four possible OpenMP pragmas to assist developers in implementing parallelization with OpenMP. 
We tackle the above-mentioned challenges by (a) generating a dataset containing 18598 parallelizable and 13972 non-parallelizable loops from benchmarks, GitHub projects, and synthetic data, (b) introducing a heterogeneous augmented-AST (aug-AST) representation for loops that considers both textual and structural information of code, and (c) training the heterogeneous aug-AST of the loops in our dataset using a heterogeneous graph neural network.

In particular, this paper makes the following contributions:
\begin{itemize}
    \item \textbf{Dataset.} OMP\_Serial: a C serial loop dataset with labels that can be used for parallelization or other code analysis purposes.

    \item \textbf{Method. }Introducing a heterogeneous augmented AST code representation suitable for parallelism detection and other downstream tasks. 

    \item \textbf{Evaluation. } Comparing the proposed graph-based approach with AST and token-based code representation approach. 

    \item \textbf{Application. } Implementing a heterogeneous GNN on the proposed dataset and comparing the results with state-of-the-art parallelization tools.
\end{itemize}
\section{Motivation Examples}
\label{motivation}


This section demonstrates and discusses the limitations of three widely used algorithm-based auto-parallelization tools: DiscoPoP \cite{li2015discopop}, Pluto \cite{pluto}, and autoPar \cite{quinlan2011rose}. These non-ML tools are generally classified into static and dynamic (hybrid) approaches. 


Dynamic or hybrid parallelization tools like DiscoPoP \cite{li2015discopop} identify parallelism with runtime dynamic information generated by executing the programs.  Profiling and executing programs are costly in terms of time and memory. In contrast, static analysis tools such as Pluto \cite{pluto} and autoPar \cite{quinlan2011rose} examine source codes statically without execution. However, these static analysis tools tend to be overly conservative, often overlooking parallelization opportunities. In addition to their inherent limitation, the use of non-ML tools is constrained due to their need for compilation or execution of the program. When applied to the OMP\_Serial dataset introduced in section \ref{sec:data}, only 10.3\% and 3.7\% of the C loops can be processed with autoPar (static) and DiscoPoP (dynamic), respectively.

\begin{figure}[!ht]
  \begin{center}
     \includegraphics[width=0.47\textwidth]{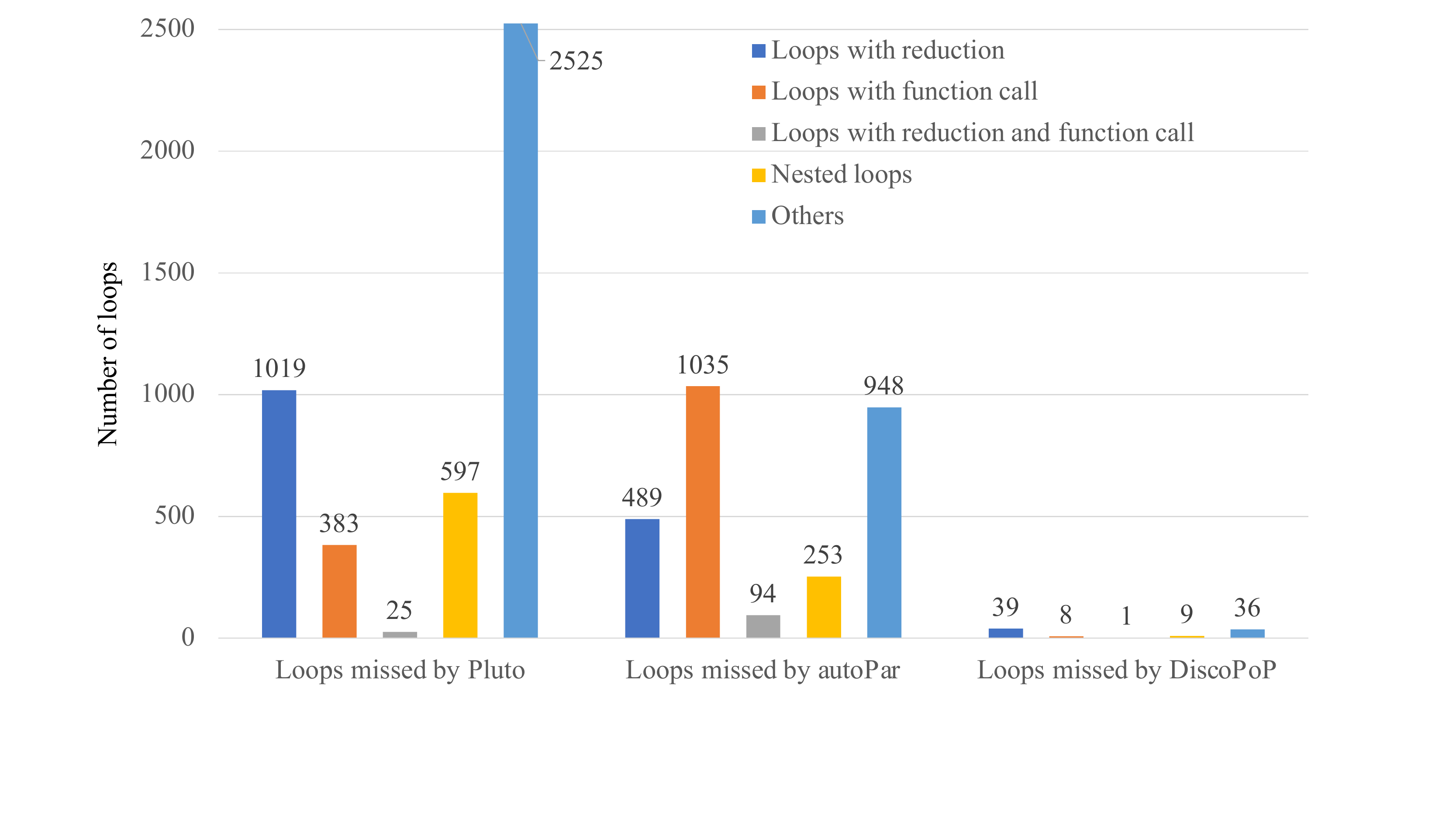}
     \caption{Category-wise loops missed by renowned parallelization assistant tools. The results are generated using the OMP\_Serial dataset introduced in section \ref{sec:data}.}{}
    \label{fig:mistake-categorized}
  \end{center}
\end{figure}

There are four types of loops where tools mostly make mistakes in our observation: loops with reduction, loops with function calls, loops with reduction and function calls, and nested loops. Listings \ref{lst:parallel_missed_plt_atp_1}, \ref{lst:parallel_missed_plt}, \ref{lst:parallel_missed_atp}, \ref{lst:parallel_missed_dp} and \ref{lst:nest_parallel_missed_dp_plt} present the example of mistakes made by autoPar, Pluto, and DiscoPoP. 
Figure \ref{fig:mistake-categorized} illustrates the statistic of our findings regarding the number and type of the loops these tools fail to detect parallelism.

\begin{minipage}{\linewidth}
\begin{lstlisting}[frame=single, caption={Parallel loop with reduction and function call. DiscoPoP, Pluto, and autoPar fail to detect the parallelism due to the $fabs$ function call.}, label={lst:parallel_missed_plt_atp_1}, captionpos=b, basicstyle=\scriptsize, language=C]
  for (i = 0; i < 30000000; i++)
    error = error + fabs(a[i] - a[i + 1]);
\end{lstlisting}
\end{minipage}

\begin{minipage}{\linewidth}
\begin{lstlisting}[frame=single, caption={Parallel loop with reduction and function call missed by Pluto because of the $abs$ function call.},label={lst:parallel_missed_plt}, captionpos=b, basicstyle=\scriptsize, language=C]
  for (int i = 0; i < num_pixels; i++) {
    fitness += (abs(objetivo[i].r - 
    individuo->imagen[i].r) + 
    abs(objetivo[i].g - 
    individuo->imagen[i].g)) + 
    abs(objetivo[i].b - 
    individuo->imagen[i].b);
  }
\end{lstlisting}
\end{minipage}

\begin{minipage}{\linewidth}
\begin{lstlisting}[frame=single, caption={Parallel loop with a function call missed by autoPar because of the $square$ function call.},label={lst:parallel_missed_atp}, captionpos=b, basicstyle=\scriptsize, language=C]
  for (int i = 0; i < size; i++) {
    vector[i] = square(vector[i]);
  }
  float square(int x) {
    int k = 0;
    while (k < 5000)
      k++;
    return sqrt(x);
  }
\end{lstlisting}
\end{minipage}

\begin{minipage}{\linewidth}
\begin{lstlisting}[frame=single, caption={Parallel loop with reduction missed by Discopop because of the reduction operation on variable $v$.},label={lst:parallel_missed_dp}, captionpos=b, basicstyle=\scriptsize, language=C]
  for (int i=0; i<N; i+= step) {
      v += 2;
      v = v + step; 
  }
\end{lstlisting}
\end{minipage}

\begin{minipage}{\linewidth}
\begin{lstlisting}[frame=single, caption={Nested parallel loop (outermost for) missed by Discopop and Pluto.},label={lst:nest_parallel_missed_dp_plt}, captionpos=b, basicstyle=\scriptsize, language=C]
  for (j = 0; j < 4; j++)
    for (i = 0; i < 5; i++)
      for (k = 0; k < 6; k += 2)
        l++;
\end{lstlisting}
\end{minipage}

We are motivated to explore cutting-edge machine learning techniques for a more feasible and precise solution. The evaluation in section \ref{sec:res} demonstrates that our proposed approach surpasses the tools we examined in detecting parallelism within complex-structure loops.

\section{Background}
\label{sec:bkg}

The field of source code analysis encompasses a broad spectrum of topics, including bug detection, optimization, and auto-parallelization. Specifically, the parallelization of sequential programs constitutes a sub-field that concentrates on tasks such as detecting parallelism, classifying parallelization patterns, and implementing parallelization. This section delves into the background of parallelization analysis and explores machine learning approaches pertinent to this task.

\subsection{Auto-parallelization and Algorithm-based Tools}
Sequential program parallelization poses considerable challenges, generally involving two phases: parallelism identification and parallelization implementation. Parallelism identification entails the analysis of sequential program fragments to identify opportunities for parallelism. Parallelization implementation or execution involves capitalizing on the detected parallelism to fully exploit the hardware capabilities.

Parallelism can be expressed through two fundamental concepts: task-level parallelism and loop-level parallelism. 
Task-level parallelism demarcates regions within an application that can be executed simultaneously on multiple cores or threads. Task-level parallelism methods require predefined distinct regions in the program, which can limit fine-grained opportunities. Loop-level parallelism considers loop bodies parallel regions, where iterations can be distributed across threads \cite{Wismuller2011}. This work primarily focuses on loop-level parallelism.

The identification of loops eligible for parallelism often relies on the program author, as modern compilers are unable to fully take advantage of parallel loop classification. However, This process imposes a significant burden on developers, particularly for extensive projects. Most dynamic approaches employ dependency analysis to record execution order constraints between instructions, enabling a more accurate automatic parallelizable loop identification. In contrast, static methods infer dependencies by conservatively analyzing the program during compilation.

Different static, dynamic, and hybrid (i.e., combining static and dynamic) tools have been developed to automatically identify parallelization opportunities. Polly \cite{grosser2012polly}, an automatic parallelism detection tool, is based on static analysis, LLVM \cite{lattner2004llvm}, and the polyhedral model. Kremlin \cite{garcia2012kremlin} determines the critical path length within the loops using dependency information and subsequently calculates a metric, namely self-parallelism, for parallelism detection. Alchemist \cite{zhang2009alchemist} identifies parallelization candidates by comparing the number of instructions with the read-after-write (RAW) dependencies, both of which are generated by Valgrind \cite{nethercote2007valgrind} during runtime. DiscoPoP \cite{li2015discopop, new1:Huda:2016:parallel_patterns} extracts dynamic profiling and instruction dependency data from instrumented sequential programs. Information like dependency type, the number of incoming and outgoing dependencies, and critical path length are extracted from a data dependency graph for parallelism detection. As a hybrid method tool, DiscoPoP provides comprehensive dynamic analysis statistics that complement static analysis, yielding an improved understanding conducive to detecting parallel opportunities.

\subsection{Machine Learning-based Auto-Parallelization}
Machine learning, as defined by Alpaydin et al. \cite{alpaydin2020introduction}, involves programming computers to optimize a performance criterion using example data or past experience. Despite its potential, machine learning techniques have been under-explored and infrequently employed in parallelization analysis tasks. Fried et al. \cite{fried2013predicting} investigated an automatic method for classifying regions of sequential programs that could be parallelized, using benchmarks with hand-annotated OpenMP directives for training. 
Tournavitis et al. \cite{DBLP:conf/pldi/TournavitisWFO09} applied SVM in conjunction with static and dynamic features extracted from source codes to identify parallel regions in programs. They used NAS parallel benchmarks \cite{jin1999openmp} and SPEC OMP benchmarks \cite{aslot2001specomp} to evaluate their model. Machine learning techniques have achieved significant progress since \cite{fried2013predicting}'s and \cite{DBLP:conf/pldi/TournavitisWFO09}'s work, with the recent advancements demonstrating the capabilities of deep neural networks in code representation \cite{cummins2021programl, ma2021learning} and parallelization analysis \cite{shen2021towards, chen2022multi}.

\subsection{Code Representations}
The representation of code is crucial for applying machine learning techniques in the area of code analysis. This subsection discusses commonly used code representations and their corresponding machine learning approaches.

\noindent
\textbf{Token.} Programming tokens are fundamental elements that comprise the source code of a program.  A token is a string of characters that can be classified as constants, identifiers, operators, reserved words, or separators according to the syntax of the programming language. 
Inspired by word embedding in natural language processing (NLP), various studies have focused on generating token-based embedding that can serve as input for machine learning approaches. 
The state-of-the-art token embedding method, \textit{code2vec} \cite{alon2019code2vec}, is trained on the task of predicting method names.

\noindent
\textbf{AST.} 
The abstract syntax tree (AST) is one of the most viable representations for code. Every programming language has an explicit context-free grammar, allowing source code to be parsed into an abstract syntax tree (AST) that represents the source code's abstract syntactic structure. Each non-leaf node in an AST corresponds to a non-terminal in the context-free grammar that conveys structural information, while each leaf node corresponds to a terminal in the context-free grammar encoding program text. Figure \ref{fig:example-ast} illustrates an example of AST for listing 1. An AST can be easily converted back to source code. As our work focuses on parallelism at the loop level, we concentrate on partial ASTs that represent the desired loop.

\noindent
\textbf{CFG.} 
The control flow graph (CFG) delineates the sequence in which code statements are executed and the requirements that must be satisfied for a specific path of execution.
Nodes represent statements and predicates, while directed edges connect them and indicate the flow of control.
Although edges of CFGs need not follow any specific order, as in abstract syntax trees, it is still necessary to identify each edge as true, false, or otherwise
CFG has been employed for various purposes, such as detecting versions of well-known malicious apps and guidng fuzz testing tools.
They are also now a common code representation in reverse engineering to aid in program comprehension.
However, control flow graphs do not reveal data flow, making them unsuitable for detecting statements that process data modified by an attacker, a limitation particularly relevant to tasks like vulnerability analysis. 


\noindent
\textbf{Comprehensive graph representations.} Recent works with code representation have focused on comprehensive graph representations to incorporate more information about programs. Ben-Nun et al. \cite{ben2018neural} aimed to create an embedded representation of code based on LLVM IR, introducing an intermediate representation of programs by combining NLP techniques with code dependencies.

Cummins et al. \cite{cummins2021programl} expanded upon the work of Ben-Nun et al. to propose an IR graph representation called PrograML, which is both comprehensive and rich in code information. The downstream task experiments set a new state-of-the-art standard. However, the requirements for using PrograML are stringent due to LLVM compilation, and only 31.2\% of the data in our dataset can be processed with PrograML. Consequently, we adopt AST as our base representation of code to utilize all the data for training.

\subsection{Heterogeneous Graph Neural Networks (HGNN)}
Graph Neural Networks (GNN) models have gained success in various research domains, including biology \cite{zhang2021graph,https://doi.org/10.48550/arxiv.2203.09456}, natural language processing \cite{DBLP:journals/corr/abs-1809-05679,DBLP:journals/corr/abs-1910-02356}, image processing \cite{DBLP:journals/corr/abs-2201-12633,8954160}, and software engineering \cite{DBLP:journals/corr/abs-1711-00740,Kammoun_2022, new1:Huda:2016:parallel_patterns, tehranijamsaz2022learning}. The application of GNNs relies on the ability to represent sequential data or databases as a complex structure with large-scale nodes and edges with structural information \cite{DBLP:journals/corr/KipfW16}. However, the homogeneous representation of these GNN models hindered their ability to represent meaningful information for prediction. Heterogeneous Graph Neural Network (HGNN) models are proposed to overcome this challenge \cite{zhang2019heterogeneous}. Compared to original GNNs, HGNN has the following advantages. First, HGNNs allow nodes to connect to all types of neighborhood nodes. In HGNNs, we can define the connection between any type of node without any restriction, which overcomes the drawback of several graph datasets that restrict the type of source node and target node for each edge, such as in \cite{zhang2019heterogeneous}. Second, HGNNs can accept not only different types of nodes but also nodes with different attributes. For example, with an academic graph, HGNN allows embedding information of profile picture and description of the author, as well as embedding information of textual content of Paper node, since Paper has no information like "profile picture". HGNNs propose a new mechanism for concatenating information and linear transformation between nodes to handle this. Third, HGNNs provide a solution for aggregating neighborhood information between neighbor nodes of different types to a more meaningful embedding per each iteration of training/ inference. To achieve this, HGNN allows representing the learning thanks to different types and weights of edges beside the nodes. The first complete HGNN model was proposed by Zhang et al. \cite{zhang2019heterogeneous}, called HetGNN. Hu et al. \cite{hu2020heterogeneous} proposed HGT, a transformer-based HGNN model that utilizes the graphs' properties more efficiently than HetGNN \cite{zhang2019heterogeneous} by decomposing interaction and transformation matrices to capture common and specific patterns of relationships between nodes and edges' types. Moreover, HGT allows embedding dynamic features such as the timeline of nodes and edges. From the work of Hu et al. \cite{hu2020heterogeneous}, we justify the original HGT model to be trained and inference on parallelism detection.


\section{Dataset Selection and Analysis}
\label{sec:data}
\begin{table*}[!htb]
\centering
\caption{Statistic Summary of the proposed OMP\_Serial dataset comprises synthetic code and code collected from GitHub. Each data in the OMP\_Serial represents a loop with labels indicating whether it is parallelizable or not. Parallelizable loops also include parallel pattern labels. The Loops column displays the number of loops for each type of pragmas. The Function Call and Nested Loops columns represent the number of loops with functions and nested loops for each type of pragmas, respectively. The Avg. LOC stands for the average length of code.}
\vskip 0.15in
\begin{tabular}{|c|c|c|c|c|c|c|c|}
\hline
Source                     & Type                      & Total Loops            & Pragma Type & \begin{tabular}[c]{@{}l@{}}Loops \end{tabular} & \begin{tabular}[c]{@{}l@{}}Function Call\end{tabular} & \begin{tabular}[c]{@{}l@{}}Nested Loops\end{tabular} & Avg. LOC \\ \hline
\multirow{5}{*}{GitHub}    & \multirow{4}{*}{Parallel} & \multirow{4}{*}{18598} & reduction   & 3705             & 279                  & 887                & 6.35     \\ \cline{4-8} 
                           &                           &                        & private     & 6278             & 680                  & 2589               & 8.51     \\ \cline{4-8} 
                           &                           &                        & simd        & 3574             & 42                   & 201                & 2.65     \\ \cline{4-8} 
                           &                           &                        & target      & 2155             & 99                   & 191                & 3.04     \\ \cline{2-8} 
                           & Non-parallel              & 13972                  & -           & -                & 3043                 & 5931               & 8.59     \\ \hline
\multirow{3}{*}{Synthetic} & \multirow{2}{*}{Parallel} & \multirow{2}{*}{400}   & reduction   & 200              & 200                  & 100                & 31.59    \\ \cline{4-8} 
                           &                           &                        & private (do-all)     & 200              & 200                  & 100                & 28.26    \\ \cline{2-8} 
                           & Non-parallel              & 700                    & -           & -                & 0                    & 0                  & 6.43     \\ \hline
\end{tabular}
\label{tab:dataset_sum}
\end{table*}

In this study, we propose a dataset, OMP\_Serial, from two distinct sources: open-source projects containing OpenMP pragmas and synthetic codes with specific parallelization patterns generated by template programming. In this section, we will discuss both approaches in detail.

\subsection{Open-source code data}

Our primary source of data is GitHub, where we crawled around 16000 source files from over 6000 repositories. We focused on $C$ source files containing loops with and without OpenMP pragmas (pragmas can be either "\#pragma omp parallel for" or "\#pragma omp for"), ensuring that developers have intentionally used OpenMP directives in their code. To validate the data, we attempted to compile all the source codes using Clang to verify their correctness. Out of the 16000 source files, we were able to compile and retain 5731 source files for further analysis and experiments. Finally, we examined the label of the collected data using parallelization tools: Pluto, autoPar, and DiscoPoP and observed a small number of parallel loops missed by developers.

\subsection{Data Processing}
\label{sec:data_pre}
Data processing is necessary for the crawled source codes. The source codes are parsed to extract loops with comments removed and pragmas extracted. The loops are initially labeled as either parallel or non-parallel based on the presence of OpenMP pragmas. Loops without a pragma are classified as non-parallel. Parallel loops with OpenMP pragmas are further divided into four categories, namely $private$, $reduction$, $simd$, and $target$ based on the extracted pragma and verified with various parallelization tools.  Consequently, the OMP\_Serial dataset comprises labeled loops with their corresponding pragma clause, if present.

\subsection{Synthetic data}


To ensure pattern diversity for the OMP\_Serial dataset, we complemented the filtered crawled data with synthetic data. Both the crawled and synthetic data will be processed as described in section \ref{sec:data_pre}. We utilized \textbf{Jinja2} \cite{ronacher2008jinja2} to generate complete C programs. For the do-all and reduction patterns, we created ten templates for each pattern and generated 20 variations of C source files from each template. We sourced the templates mainly from well-known parallel benchmarks such as the NAS Parallel Benchmark \cite{jin1999openmp}, PolyBench \cite{pouchet2017polybench}, the BOTS benchmark \cite{duran2009barcelona}, and the Starbench benchmark \cite{andersch2013benchmark}. To create complete C programs, we inserted randomly generated variables, constants, and operators into the templates. The variable names were generated using a combination of English language alphabets (a-z, A-Z), digits (0-9), and underscores (\_).  For do-all loops, we considered the operators: $+,\ -,\ *,\ /$. For reduction loops, we considered only $+$ and $*$ operators since reduction operations need to be associative and commutative for parallelization.

We used DiscoPoP to verify the generated reduction and do-all templates. Loops not identified as do-all or reduction by DiscoPoP were manually checked for inter-iteration dependencies or data-race conditions. If such conditions existed in the loop body, they were labeled as non-parallel loops. More details and examples on the generation of synthetic data can be found in Appendix \ref{app:synthetic}. Finally, the OMP\_Serial dataset, comprising both open-source and synthetic data, is summarized in Table \ref{tab:dataset_sum}.

\section{Approach}
\label{sec:approach}
The representation of code is crucial for any analysis task. We propose an augmented heterogeneous AST representation for comprehending code in semantic and structural views. We first introduce the augmented AST (aug-AST) representation based on the control flow graph (CFG) and token distance in text format. Next, we append the types of nodes and edges in the aug-AST and build the augmented heterogeneous AST graph for each data point in our OMP\_serial dataset. We use the heterogeneous graph transformer (HGT) model \cite{hu2020heterogeneous} as our base model, taking the augmented heterogeneous AST graph as input.

\begin{figure*}[h]
    \centering
    \includegraphics[width=16cm]{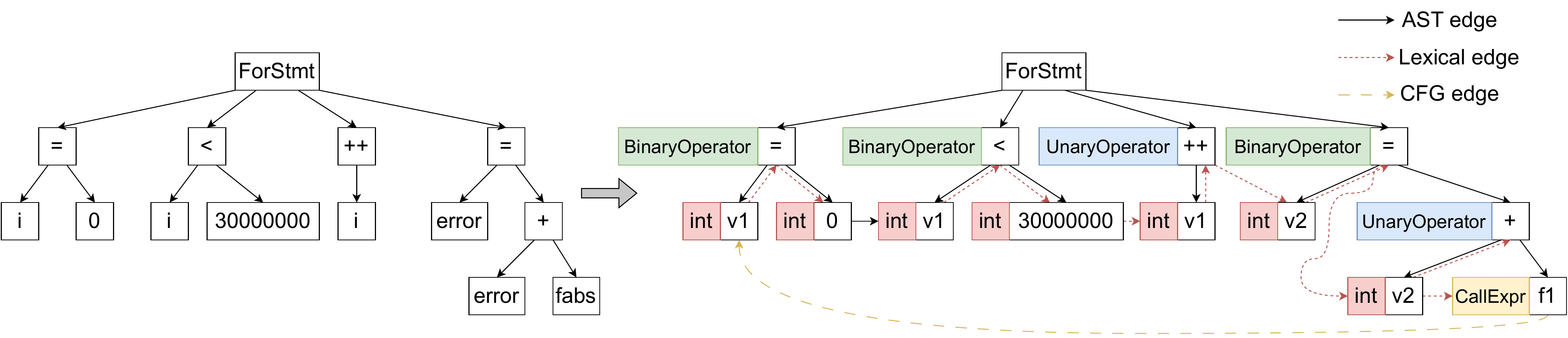}
    \caption{An example of the proposed heterogeneous augmented AST (Heterogeneous aug-AST) representation of code in Listing 1 is shown. The colored blocks indicate the heterogeneous attributes assigned to the AST nodes. The red and yellow lines represent the control flow graph (CFG) and token representation, respectively.}
    \label{fig:example-ast}
\end{figure*}

\subsection{Code representation}
Code representations like AST and CFG provide crucial data for code analysis. However, a single representation is often insufficient to capture all the dependencies and parallelism. To address this issue, we propose an augmented AST that merges edges and nodes from the CFG, creating a single graph that incorporates the benefits of each distinct representation. Additionally, we address long-dependence problems by incorporating texture edges that follow the token distance map.


\subsubsection{Transforming the Abstract Syntax Tree}

To build a joint representation, we propose an augmented AST that incorporates both AST and CFG. We express the AST as a heterogeneous graph $HA = (V_A, E_A, \lambda _A, \mu_A)$, where nodes $V_A$ represent AST tree nodes and edges $E_A$ represent corresponding tree edges labeled as AST edges by the labeling function $\lambda_A$. Each node is assigned an attribute using $\mu_A$ that corresponds to the operator or operand the node represents. Furthermore, we assign an attribute to each node to reflect the tree's ordered structure (left or right). The color blocks in Figure \ref{fig:example-ast} represent the heterogeneous node attributes, while the black edges represent edges from the AST.

\subsubsection{Merging the Control Flow Graph}

To include the CFG in the joint representation, we express it as a heterogeneous graph $GC = (V_C, E_C, \lambda_C, \cdot )$. The nodes $V_C$ represent statements and predicates in the loop AST. We also introduce edges from nodes shared by the AST and CFG to nodes in the AST graph. These edges are represented by yellow dash lines in figure \ref{fig:example-ast}, where node $f1$ is a function call node shared by both AST and CFG. These edges enable the machine learning model to identify potential data races within the function call and explore parallelization opportunities.

\subsubsection{Texture token relations}
In the work of \cite{zugner2021language}, Zugner et. al revealed that AST alone may miss important lexical token distance information, leading to difficulties in capturing long-distance dependence relations. To address this issue, we add extra edges to link each leaf with its neighbors in the token representation as shown in figure \ref{fig:example-ast}. The added lexical edges (represented by red dashes) help aug-AST track the token distance.

\subsection{Heterogeneous Graph Transformer}
In this study, the input for the Heterogeneous Graph Transformer (HGT) model is the aug-AST graph generated from the original AST plus augmented nodes and edges. An aug-AST graph is represented as a heterogeneous graph, denoted by $G=(V, E, A, R)$. Here, $V$ denotes the set of nodes, $E$ denotes the set of edges, $A$ represents the possible types of nodes in $V$, and $R$ represents the possible types of edges in $E$. For a given edge $e=(s,t)$ with source node $s$ and target node $t$, a meta-relation of the edge $e$ is defined by the type of $s$, the type of $t$, and the type of edge $e$. In our work, three types of edges are considered: parent-child edges generated by the original AST and augmented CFG and lexical edges added to capture the control flow information and the relationship between neighbor leaf nodes. In the original GNN model, information is updated from the $(l-1)$-th layer to the $l$-th layer by the formula \ref{eq:GNNLayer}.
\begin{equation}
H^{l}[t]=\text{Aggregate}(\text{Extract}(H^{l-1}[s];H^{l-1}[t];e))
    \label{eq:GNNLayer}
\end{equation}

Here, $h_v^{(l)}$ is the feature representation of node $v$ at the $l$-th layer, $\sigma$ is the activation function, $N_{r}^{out}(v)$ is the set of nodes that have an outgoing edge of type $r$ from $v$, $W_r^{(l)}$ is the trainable weight matrix for edge type $r$ at layer $l$, and $d^{(l-1)}_v$ is the degree of node $v$ in the $(l-1)$-th layer.

In formula \ref{eq:GNNLayer}, the \textit{Extract} operator extracts information from neighbor nodes $s$ to target node $t$ and the \textit{Aggregate()} combines information from all the source that has the target node as $t$. In HGT, the mechanism of passing information between layers is split into three components: Heterogeneous Mutual Attention, Heterogeneous Message Passing, and Target Specific Aggregation.

\textbf{Mutual Attention.} The input of this step is the node $t$ and a set of $N(t)$, which represents all the source nodes of the relation $r$. The heterogeneous mutual attention mechanism is calculated by taking the dot product between the source node $s$ (Key vector) and the node $t$ (Query vector). Next, the Key vector is projected using a linear projection to $h$ attention heads, where each head is represented by a vector with $\frac{d}{h}$ dimension. Similarly, the Query vector is also projected into $h$ Query vectors. For each head $h$, the Query vector is compared with the projection of the Key vector using a distinct edge-based matrix $W^{\textit{ATT}}$. Finally, the attention vector for each pair of nodes is produced by concatenating the $h$ attention heads. The gathering of all attention vectors from the set of neighbor nodes $N(t)$ to the target node $t$ is shown in the formula \ref{eq:HMA}.

\begin{equation}
\small
    \text{Attention}_{\textit{HGT}}(s,e,t) = \underset{\forall s \in N(t)}{\text{Softmax}}( \underset{i\in [1,h]}{||} ATT \\ -head^{i}(s,e,t) )
\label{eq:HMA}
\end{equation}

\textbf{Message Parsing.} While the Mutual Attention compares between Key vector and Query vector as target node and source node, the Message Passing mechanism operates in parallel. The input of Message Passing is not only the edge but also its meta relations. The formula of the Message operator is shown in the formula \ref{eq:HMP}, where the MSG-head function is calculated by a number of components.
\begin{equation}
\small
\text{Message}_{\textit{HGT}}(s,e,t)= \underset{i\in [1,h]}{||} \textit{MSG}-\textit{head}^{i}(s,e,t) 
\label{eq:HMP}
\end{equation}

The amount of components in the equation \ref{eq:HMP} is equal to the number of hidden layers. Similar to Formula \ref{eq:HMA}, the Message Passing step also needs a matrix $W^{\textit{MSG}}$ that embeds information of the edge dependency. 

\textbf{Target Specific Aggregation.} This Target Specific Aggregation operator combines the Attention operator calculated by Formula \ref{eq:HMA} and the Message operator calculated by Formula \ref{eq:HMP} to generate an update vector for each head, as shown in equation \ref{eq:aggregation}.
\begin{equation}
\small
    \bar{H}^{(l)}[t]=Aggregate(Attention(s,e,t).Message(s,e,t))
    \label{eq:aggregation}
\end{equation}
In the final step, the output of each head calculated by the formula \ref{eq:aggregation} is combined with a type-specific distribution of target node $t$ through a linear projection:
\begin{equation}
\small
    \bar{H}^{(l)}[t]=A-Linear_{(type(t))}(\sigma (\bar{H}^{(l)}[t]))+H^{(l-1)}[t]
    \label{eq:tsa}
\end{equation}
In Graph2Par, the distribution of $type(t)$ is the set of different node types in the aug-AST. In the work of Hu et al. \cite{hu2020heterogeneous}, they provide Inductive Timestamp Assignment and  Relative Temporal Encoding to represent the dynamic heterogeneous graphs. However, since Graph2Par works with static and structural information of AST, we set the same temporal encoding mechanism and deactivated the inductive timestamp assignment in our HGT model.

\section{Results}
\label{sec:res}
In this section, we present the results of our experiments aimed at answering two research questions: 1. evaluating the performance of the proposed Heterogeneous augmented AST code representation, and 2. assessing the effectiveness of the proposed Graph2Par method for OpenMP pragma suggestion. Additional training results are provided in the appendix (see Appendix \ref{app:eval}).

\subsection{Performance of the Heterogeneous aug-AST}
We demonstrate that our proposed Heterogeneous aug-AST representation outperforms both token-based and original AST representations by evaluating its performance in predicting parallelism. We compare the vanilla AST and the Heterogeneous aug-AST by using them as inputs to the same HGT model.
Additionally, we reproduce PragFormer, the work of Harel et. al \cite{harel2022learning}, to compare the performance of token representation and Heterogeneous aug-AST representation. PragFormer uses token-based representation as input to a transformer model for parallelism detection. Table \ref{tab:res_1} shows that our Heterogeneous aug-AST outperforms PragFormer in parallelism detection.

\begin{table}[!ht]
    \centering
    \caption{Result of pragma existence prediction. PragFormer uses token representations.}
    \vskip 0.15in
    \begin{tabular}{|c|c|c|c|c|}
    \hline
        ~ & Precision & Recall & F1 & Accuracy \\ \hline
        AST & 0.74 & 0.73 & 0.74 & 0.74 \\ \hline
        PragFormer & 0.81 & 0.81 & 0.80 & 0.80 \\ \hline
        Graph2Par & 0.92 & 0.82 & 0.87 & 0.85 \\ \hline
    \end{tabular}
    \label{tab:res_1}
\end{table}

\begin{table}[!ht]
    \centering
    \caption{Number of detected parallel loops comparing with algorithm-based approaches.}
    \vskip 0.15in
    \begin{tabular}{|c|c|}
    \hline
        Approach & \# of detected parallel loops \\ \hline
        Graph2Par & 17563 \\ \hline
        HGT-AST & 16236 \\ \hline
        DiscoPoP & 953 \\ \hline
        PLUTO & 1759 \\ \hline
        autoPar & 6391 \\ \hline
    \end{tabular}
    \label{tab:num_detected_al}
\end{table}

\subsection{Parallelism Discovery: Comparing with other tools}

\begin{table*}[!ht]
\centering
\caption{Comparing Graph2Par model with PLUTO, autoPar and DiscoPoP for the task of parallelism detection (Detecting the presence of "\#pragma omp for" or "\#pragma
omp parallel for")}
\vskip 0.15in
\begin{tabular}{|cc|c|c|c|c|c|c|c|c|}
\hline
\multicolumn{2}{|c|}{}                                     & TP   & TN   & FP  & FN   & Precision & Recall & F1    & Accuracy(\%) \\ \hline
\multicolumn{1}{|c|}{\multirow{2}{*}{Subset PLUTO}} & PLUTO    & 1593 & 0    & 0   & 2439 & 100.00     & 39.51  & 56.64 & 39.51        \\ \cline{2-10} 
\multicolumn{1}{|c|}{}                          & Graph2Par    & 2860 & 617    & 356   & 199 & 88.93    & 93.49  & 91.16 & 86.24        \\ \hline
\multicolumn{1}{|c|}{\multirow{2}{*}{Subset autoPar}} & autoPar  & 345  & 952  & 0 & 2059 & 100.00     & 14.35   & 25.10 & 38.65        \\ \cline{2-10} 
\multicolumn{1}{|c|}{}                          & Graph2Par    & 1800 & 897 & 187  & 472  & 90.59     & 79.23  & 84.53 & 80.36        \\ \hline
\multicolumn{1}{|c|}{\multirow{2}{*}{Subset DiscoPoP}} & DiscoPoP & 541  & 240  & 0 & 445  & 100.00     & 54.87  & 70.86 & 63.70        \\ \cline{2-10} 
\multicolumn{1}{|c|}{}                          & Graph2Par    & 635  & 366  & 64 & 161  & 90.84     & 79.77  & 84.95 & 81.65        \\ \hline
\end{tabular}
\label{tab:compare_with_tools}
\end{table*}
The results of the above experiments demonstrate that the proposed Heterogeneous aug-AST representation outperforms both original AST and token-based representations in parallelism detection.
In this subsection, we continue the evaluation of the aug-AST presentation by comparing it with well-known algorithm-based parallelism assistant tools: PLUTO, autoPar, and DiscoPoP. PLUTO and autoPar are algorithm-based static analysis tools, whereas DiscoPoP is an algorithm-based dynamic analysis tool. All three auto-parallelization tools can detect parallelism in codes they can handle.  However, parallelization pattern classification is not supported by all the tools. For example, $simd$ and $target$ clause predictions are not supported by any tools at present. 

Therefore, we conduct a performance comparison for the task of parallelism detection. As mentioned in section \ref{sec:data}, loops in the OMP\_serial dataset are labeled 1 when the OpenMP clauses are present and labeled 0 otherwise. Graph2Par predicts the parallelism within a loop by a binary classification. PLUTO directly reports the parallelism detection results within a loop. autoPar injects OpenMP clauses like "\#pragma omp parallel for" including "private" clause and "reduction" clause to the programs. We mark the detection results as parallel when the injected clauses are present. DiscoPoP can detect reduction and do-all patterns within a loop, and we considered the loops detected as either do-all or reduction by DiscoPoP as parallel loops. 

Different tools usually work with different sizes of data because they may require different information about the codes. DiscoPoP, for example, requires execution information for analysis, making it works with a much smaller dataset compared with static tools like PLUTO. Therefore, we divided our test dataset into three subsets for a fair comparison between Graph2Par and different tools. The results are presented in Table \ref{tab:compare_with_tools}. Our Graph2Par model achieves superior performance compared to the other tools, indicating its effectiveness in detecting parallelism in sequential programs.

\begin{itemize}
    \item \textbf{Subset PLUTO:} This subset contains the loops that are in our testing set and can also be successfully processed by PLUTO. This set contains 4032 loops.
    \item \textbf{Subset autoPar:} This subset contains the loops that are in our testing set and can also be successfully processed by autoPar. This set contains 3356 loops.
    \item \textbf{Subset DiscoPoP:} This subset contains the source files that are in our testing set and can also be successfully processed by DiscoPoP. This set contains 1226 loops.
\end{itemize}
We train our Graph2Par approach the three subset described above separately for comparison. In each training, one of the subsets was excluded to ensure that the model had not seen the samples before. The results are presented in tables \ref{tab:num_detected_al} and \ref{tab:compare_with_tools}. For all three subsets, our Graph2Par model achieved better precision, recall, F1 score, and accuracy than all the other tools.

\subsection{OpenMP Clause Classification}
The above results demonstrate that Graph2Par has the ability to learn the latent features of code for parallelism detection. In this subsection, we evaluate the extensibility of our Graph2Par model for predicting OpenMP pragmas, including "private", "reduction", "simd", and "target". We apply the same labeling strategy as the parallelism detection task, where the presence of the corresponding pragma determines the label of the loop. We train Graph2Par on the entire OMP\_serial dataset and evaluate on a separate test set. The results are presented in Table \ref{tab:clause_all}. We observe that our Graph2Par model performs well for the "private" and "reduction" pragma prediction tasks but struggles with the "simd" and "target" pragma prediction tasks. This is due to the limited representation of the aug-AST for certain pragma patterns, as some patterns may require additional information beyond the control flow graph and lexical edges represented by the aug-AST.  

It is worth noting that algorithm-based tools are not able to predict all of these pragmas or process every data point in our dataset. As the state-of-the-art model, PragFormer is used as a baseline for comparing the results of Graph2Par. Table \ref{tab:clause_all} shows that our Graph2Par approach outperforms the SOTA token-based approach in both "private" and "reduction" pragma prediction tasks. Overall, the results demonstrate that our Graph2Par model has the potential to be extended to other OpenMP pragma prediction tasks, but additional features and representations may be required to handle more complex patterns.

\begin{table}[h]
    \centering
    \caption{Performance of Graph2Par for four pragma prediction.}
    \vskip 0.15in
    \resizebox{0.48\textwidth}{!}{%
    \begin{tabular}{|c|c|c|c|c|c|}
    \hline
        Pragma & Approach & Precision & Recall & F1-score & Accuracy \\ \hline
        private & Graph2Par & 0.88 & 0.87 & 0.87 & 0.89 \\ \hline
        ~ & PragFormer & 0.86 & 0.85 & 0.86 & 0.85 \\ \hline
        reduction & Graph2Par & 0.9 & 0.89 & 0.91 & 0.91 \\ \hline
        ~ & PragFormer & 0.89 & 0.87 & 0.87 & 0.87 \\ \hline
        SIMD & Graph2Par & 0.79 & 0.76 & 0.77 & 0.77 \\ \hline
        ~ & PragFormer & N/A & N/A & N/A & N/A \\ \hline
        target & Graph2Par & 0.75 & 0.74 & 0.74 & 0.74 \\ \hline
        ~ & PragFormer & N/A & N/A & N/A & N/A\\ \hline
    \end{tabular}}
    \label{tab:clause_all}
\end{table}

\subsection{Dealing with False Positives}

From Table \ref{tab:compare_with_tools}, it can be observed that our proposed Graph2Par has some false positives, meaning that it predicted some loops that are not parallel as parallel loops. In contrast, traditional tools like PLUTO, autoPar, and DiscoPoP have zero false positives.  However, Graph2Par is able to detect 1.8x, 5.2x, and 1.2x more parallel loops (true positives) in the Subset PLUTO, Subset autoPar, and Subset DiscoPoP datasets, respectively. This suggests that although Graph2Par may wrongly predict some loops as parallel, it can discover more parallelization opportunities than traditional approaches that are often conservative and may miss out on such opportunities. False positives are inevitable when embracing machine learning techniques since no model is perfect and can make mistakes. Parallelizing serial programs is complex, which makes it hard to do end-to-end auto-parallelization, even with algorithm-based tools. There is more to consider for end-to-end approaches other than the parallelization pattern within the code, such as the characteristics of the platform on which the code executes, as well as the input data size and data dependencies. These factors can significantly impact the performance of the parallelized code, and their consideration is crucial for achieving optimal speedup. Therefore, it is important to carefully analyze and tune these factors in addition to identifying the parallelism opportunities within the code.
Therefore, Graph2Par handles the false positives by only providing suggestions instead of generating end-to-end parallel code. The suggestion provided by Graph2Par includes whether parallelism exists within a loop and whether the loop inhibits any parallel patterns when parallelism is present. Developers can then use this information to parallelize the loops using any framework they prefer. For example, if a developer finds that a loop is parallel and has a reduction pattern, they can easily parallelize the loop using the "\#pragma omp parallel for reduction" clause of OpenMP. However, there may be scenarios where the false positives are significant and need to be reduced to avoid confusion and save developers time. In such cases, developers may use additional tools to manually verify the suggested parallelism by Graph2Par.


\subsection{Overhead.}
When generating the proposed aug-AST representation for a loop, the steps mentioned in section \ref{sec:approach} are followed. The overhead of creating an aug-AST comes from two steps: code compilation with Clang and AST traversal with tree-sitter \cite{tree-sitter}. However, both steps introduce minimal overhead. It is important to note that the overhead of creating an aug-AST may increase for larger size codes. However, for the loops in the OMP\_serial dataset, which have an average size of 6.9 lines, the overhead is minimal and in the order of milliseconds.

\subsection{Case Study}

In the evaluation, it is observed that our proposed model can successfully identify 48 parallel loops missed by all three algorithm-based tools.  An example of one such loop is presented in Listing \ref{lst:parallel_missed_all_1}, and other examples can be found in Listings \ref{lst:parallel_missed_plt_atp_1}, \ref{lst:parallel_missed_plt}, \ref{lst:parallel_missed_atp}, \ref{lst:parallel_missed_dp}, and \ref{lst:nest_parallel_missed_dp_plt} in the motivation examples. These results demonstrate the effectiveness of our Graph2Par approach in detecting parallelism opportunities that are missed by traditional algorithm-based tools.



\begin{minipage}[H]{\linewidth}
\begin{lstlisting}[frame=single, caption={Parallel loop missed by DiscoPoP, PLUTO and autoPar with array and reduction}, label={lst:parallel_missed_all_1}, captionpos=b, basicstyle=\scriptsize, language=C]
  for (i = 0; i < 1000; i++){
    a[i] = i * 2;
    sum += i;
  }
\end{lstlisting}
\end{minipage}

Another example is shown in Listing \ref{lst:parallel_missed_all_2}. We believe that the conservative nature of non-AI-based parallelism assistant tools may be the reason for missing such opportunities. In this specific example, although there is a reduction operation on the variable $sum$ and memory access to the 2D array $a$, only the $j$ index is changing, and there are no inter-iteration dependencies. Therefore, this loop can be executed in parallel, and it is successfully detected by our Graph2Par model.

\begin{minipage}[H]{\linewidth}
\begin{lstlisting}[frame=single, caption={Parallel loop missed by DiscoPoP, PLUTO and autoPar with array and reduction}, label={lst:parallel_missed_all_2}, captionpos=b, basicstyle=\scriptsize, language=C]
  for (j = 0; j < 1000; j++){
    sum += a[i][j] * v[j];
  }
\end{lstlisting}
\end{minipage}

Furthermore, our proposed Graph2Par model can handle parallelism detection in nested loops effectively, which is a challenging problem due to the complex dependencies between the loops. As an example, in Listing \ref{lst:parallel_missed_all_4}, the outer parallel loop has been missed by all traditional parallelism assistant tools due to its nested structure. However, our model successfully detects that the outer-most $for$ loop can be parallelized. By observing that each cell of the 3-d array $a$ will eventually have the same value and that $m$ is just a constant, we can verify that there are no loop-carried dependencies, and the loop can be safely parallelized.


\begin{minipage}[H]{\linewidth}
\begin{lstlisting}[frame=single, caption={Parallel loop missed by DiscoPoP, PLUTO and autoPar with nested loop}, label={lst:parallel_missed_all_4}, captionpos=b, basicstyle=\scriptsize, language=C]
  for (i = 0; i < 12; i++) {
    for (j = 0; j < 12; j++) {
      for (k = 0; k < 12; k++) {
        tmp1 = 6.0 / m;
        a[i][j][k] = tmp1 + 4;
      }
    }
  }
\end{lstlisting}
\end{minipage}

\section{Related Work}

Recent research has shown an increasing trend in employing machine learning techniques for parallelization analysis. These studies can be broadly classified into two categories based on their code representations. Token-based code analysis studies \cite{fried2013predicting, harel2022learning} used natural language processing (NLP) models trained on raw code text data. In contrast, recent studies such as \cite{shen2021towards, chen2022multi} have leveraged structured graphical models with the structural representation of code, such as the Abstract Syntax Tree (AST). Compared to these works, our proposed Heterogeneous augment-AST representation is easy to process and contains rich information on nodes and edges, enabling more accurate and efficient parallelization analysis.

\section{CONCLUSION}
In this paper, we propose a static approach to discover parallelism in sequential programs using an augmented AST representation. To address the issue of data insufficiency, we created the OMP\_Serial dataset, which can be used for other parallelization tasks as well. We evaluate the aug-AST representation using a GNN-based model, and it outperforms traditional parallelization tools as well as token-based machine learning approaches. However, there is still room for improvement in our model. Currently, Graph2Par can only detect whether a pragma is applicable for a loop or not, but future research directions could focus on developing a model that can generate complete OpenMP pragmas for sequential loops.


\bibliography{mlsys23_main}
\bibliographystyle{mlsys2023}

\clearpage

\end{document}